\documentclass[sigconf, nonacm]{acmart}
\pdfoutput=1
\newcommand\vldbdoi{10.14778/3447689.3447713}
\newcommand\vldbpages{1102 - 1110}
\newcommand\vldbvolume{14}
\newcommand\vldbissue{6}
\newcommand\vldbyear{2021}
\newcommand\vldbauthors{\authors}
\newcommand\vldbtitle{\shorttitle} 
\newcommand\vldbavailabilityurl{https://github.com/xgfs/frede}
\newcommand\vldbpagestyle{empty} 

\usepackage[utf8]{inputenc}
\usepackage[T1]{fontenc}
\usepackage{xcolor}
\usepackage{colortbl}
\usepackage{url}
\usepackage{tabularx}
\usepackage{booktabs}
\usepackage{amsfonts}
\usepackage{amsmath}
\usepackage{amsthm}
\usepackage{pifont}
\usepackage{xspace}
\usepackage{xfrac} \usepackage{tikz}
\usepackage{pgfplots}
\usetikzlibrary{fit,positioning,calc,shapes,backgrounds,decorations.pathmorphing}
\pgfplotsset{compat=1.14}
\usepackage{hyphenat}
\usepackage{balance}
\usepackage{multirow}
\usepackage{nicefrac}
\usepackage{enumitem}
\usepackage{algorithm}
\usepackage[noend]{algpseudocode}
\usepackage{afterpage}

\usetikzlibrary{matrix}
\usepgfplotslibrary{groupplots}

\newcommand{\spara}[1]{\smallskip\noindent{\bf #1}}
\newcommand{\mpara}[1]{\medskip\noindent{\bf #1}}
\newcommand{\para}[1]{\noindent{\bf #1}}

\definecolor{cycle1}{RGB}{0, 0, 191}
\definecolor{cycle2}{RGB}{106, 191, 0}
\definecolor{cycle3}{RGB}{191, 0, 0}
\definecolor{cycle4}{RGB}{0, 163, 191}
\definecolor{cycle5}{RGB}{191, 0, 121}
\definecolor{cycle6}{RGB}{0, 109, 191}
\definecolor{cycle7}{RGB}{191, 106, 0}
\definecolor{cycle8}{RGB}{191, 0, 63}
\definecolor{cycle9}{RGB}{141, 0, 191}
\definecolor{cycle10}{RGB}{189, 191, 0}
\definecolor{cycle11}{RGB}{0, 70, 191}
\definecolor{cycle12}{RGB}{191, 142, 0}
\definecolor{cycle13}{RGB}{191, 0, 178}
\definecolor{cycle14}{RGB}{191, 70, 0}
\definecolor{cycle15}{RGB}{0, 139, 191}
\definecolor{cycle16}{RGB}{0, 189, 191}
\definecolor{cycle17}{RGB}{191, 0, 90}
\definecolor{cycle18}{RGB}{96, 0, 191}
\definecolor{cycle19}{RGB}{191, 0, 37}
\definecolor{cycle20}{RGB}{0, 47, 191}

\newcommand{\cmark}{\textcolor{cycle2}{\ding{52}}} \newcommand{\xmark}{\textcolor{cycle3}{\ding{56}}}
\newcommand{\win}{\cellcolor{cycle2!30}}

\newcommand*{\belowrulesepcolor}[1]{\noalign{\kern-\belowrulesep
    \begingroup
      \color{#1}\hrule height\belowrulesep
    \endgroup
  }}
\newcommand*{\aboverulesepcolor}[1]{\noalign{\begingroup
      \color{#1}\hrule height\aboverulesep
    \endgroup
    \kern-\aboverulesep
  }}

\newtheorem{definition}{Definition}
\newtheorem*{proof*}{Proof}
\newtheorem{theorem}{Theorem}

\newcommand*{\ourwork}{\textsc{FREDE}\xspace}
\newcommand{\deepwalk}{\textsc{DeepWalk}\xspace}
\newcommand{\lineemb}{\textsc{LINE}\xspace}
\newcommand{\verseemb}{\textsc{VERSE}\xspace}

\newcommand{\nodetovec}{\textsc{Node}2\textsc{vec}\xspace}

\newcommand{\hoppe}{\textsc{HOPE}\xspace}
\newcommand{\arope}{\textsc{AROPE}\xspace}
\newcommand{\appr}{\textsc{ApproxPPR}\xspace}
\newcommand{\nrp}{\textsc{NRP}\xspace}
\newcommand{\netmf}{\textsc{NetMF}\xspace}
\newcommand{\netsmf}{\textsc{NetSMF}\xspace}

\newcommand{\strap}{\textsc{STRAP}\xspace}
\newcommand{\prone}{\textsc{ProNE}\xspace}
\newcommand{\nodesketch}{\textsc{NodeSketch}\xspace}
\newcommand{\louvain}{\textsc{LouvainNE}\xspace}

\newcommand{\dyoutube}{\textsc{YouTube}\xspace}
\newcommand{\dblogcatalog}{\textsc{BlogCatalog}\xspace}
\newcommand{\dvk}{\textsc{VK}\xspace}

\newcommand{\dcocit}{\textsc{CoCit}\xspace}
\newcommand{\dcoa}{\textsc{CoAuthor}\xspace}
\newcommand{\dppi}{\textsc{PPI}\xspace}
\newcommand{\dpos}{\textsc{POS}\xspace}
\newcommand{\dflickr}{\textsc{Flickr}\xspace}

\newcommand*{\bigO}{\mathcal{O}}

\newcommand{\ppr}{\mathrm{PPR}}

\newcommand{\xvec}{\mathbf{x}}
\newcommand{\rvec}{\mathbf{r}}
\newcommand{\smat}{\mathrm{S}}
\newcommand{\wmat}{\mathrm{W}}
\newcommand{\xmat}{\mathrm{X}}
\newcommand{\ymat}{\mathrm{Y}}
\newcommand{\umat}{\mathrm{U}}
\newcommand{\vmat}{\mathrm{V}}
\newcommand{\mmat}{\mathrm{M}}
\newcommand{\amat}{\mathrm{A}}
\newcommand{\pmat}{\mathrm{P}}
\newcommand{\dmat}{\mathrm{D}}

\newcommand{\adj}{\amat}
\newcommand{\emb}{\wmat}
\newcommand{\apsim}{\tilde{\smat}}

\newcommand{\reals}{\mathbb{R}}

\newcommand{\pkr}[1]{{#1}} \newcommand{\dmr}[1]{{#1}}

\newcommand{\pkq}[1]{{#1}} \newcommand{\pk}[1]{{#1}}
\newcommand{\pkn}[1]{{#1}}
\newcommand{\pknn}[1]{{#1}} \newcommand{\pkw}[1]{#1}
\newcommand{\pko}[1]{#1} \newcommand{\pki}[1]{#1}
\newcommand{\pkv}[1]{#1} \newcommand{\dmv}[1]{#1} \newcommand{\atv}[1]{#1}

\pgfplotsset{
  /pgfplots/line legend with two lines/.style 2 args={
    legend image code/.code={
      \draw[mark phase=2, mark repeat=2, #1] plot coordinates {(0cm, -0.1cm) (0.25cm, -0.1cm) (0.5cm, -0.1cm)};
      \draw[mark phase=2, mark repeat=2, #2] plot coordinates {(0cm, 0.1cm) (0.25cm, 0.1cm) (0.5cm, 0.1cm)};
    }
  }
}

\definecolor{darkgreen}{rgb}{0.0, 0.4, 0.26}

\begin{document} \title{FREDE: Anytime Graph Embeddings}
\author{Anton Tsitsulin}
\affiliation{\institution{University of Bonn}
}
\email{tsitsulin@bit.uni-bonn.de}
\author{Marina Munkhoeva}
\affiliation{\institution{Skoltech}
}
\email{marina.munkhoeva@skoltech.ru}
\author{Davide Mottin}
\affiliation{\institution{Aarhus University}
}
\email{davide@cs.au.dk}
\author{Panagiotis Karras}
\affiliation{\institution{Aarhus University}
}
\email{panos@cs.au.dk}
\author{Ivan Oseledets}
\affiliation{\institution{Skoltech}
}
\email{ivan.oseledets@skoltech.ru}
\author{Emmanuel M\"uller}
\affiliation{\institution{TU Dortmund}
} 
\email{mueller@bit.uni-bonn.de}

\begin{abstract}
Low-dimensional representations, or \emph{embeddings}, of a graph's nodes \pkv{facilitate several practical data science and data engineering tasks. As such embeddings} \pkw{rely, explicitly or implicitly, on a similarity measure} among nodes, \pkv{they require the computation of a quadratic similarity matrix, inducing} \pkw{a tradeoff between space complexity and embedding quality. \pkv{To date, no graph embedding work combines (i) linear space complexity, (ii) a nonlinear transform as its basis, and (iii) nontrivial quality guarantees.}}
In this paper we \pki{introduce \ourwork (\emph{FREquent Directions Embedding}), \pkv{a graph embedding based on matrix sketching that combines those three desiderata.}} \pki{Starting out from the observation} \pkw{that embedding methods \pk{aim to} preserve the covariance among the rows of a similarity matrix}, \ourwork{} \pkw{iteratively improves on quality while individually processing rows of \pkv{a nonlinearly transformed PPR similarity matrix derived from a state-of-the-art graph embedding method} and provides, \emph{at any iteration}, column-covariance approximation guarantees \pkn{in due course almost indistinguishable from those of the optimal approximation by SVD}.} Our experimental evaluation on variably sized networks shows that \ourwork performs almost as well as SVD \pkw{and competitively against state\hyp{}of\hyp{}the\hyp{}art embedding methods in diverse data science tasks, even when it is based on as little as 10\% of node similarities.}
\end{abstract} \maketitle

\pagestyle{\vldbpagestyle}
\begingroup\small\noindent\raggedright\textbf{PVLDB Reference Format:}\\
\vldbauthors. \vldbtitle. PVLDB, \vldbvolume(\vldbissue): \vldbpages, \vldbyear.\\
\href{https://doi.org/\vldbdoi}{doi:\vldbdoi}
\endgroup
\begingroup
\renewcommand\thefootnote{}\footnote{\noindent
This work is licensed under the Creative Commons BY-NC-ND 4.0 International License. Visit \url{https://creativecommons.org/licenses/by-nc-nd/4.0/} to view a copy of this license. For any use beyond those covered by this license, obtain permission by emailing \href{mailto:info@vldb.org}{info@vldb.org}. Copyright is held by the owner/author(s). Publication rights licensed to the VLDB Endowment. \\
\raggedright Proceedings of the VLDB Endowment, Vol. \vldbvolume, No. \vldbissue\ ISSN 2150-8097. \\
\href{https://doi.org/\vldbdoi}{doi:\vldbdoi} \\
}\addtocounter{footnote}{-1}\endgroup

\ifdefempty{\vldbavailabilityurl}{}{
\vspace{.3cm}
\begingroup\small\noindent\raggedright\textbf{PVLDB Artifact Availability:}\\
The source code, data, and/or other artifacts have been made available at \url{\vldbavailabilityurl}.
\endgroup
}

\begin{figure}[t]
        \begin{tikzpicture}
        	\begin{axis}[
	    scale only axis,
		ylabel={\% of nodes visited},
		ymin=0,
		ymax=100,
		height=3cm,
		width=0.7\linewidth,
		xtick=\empty,
		xmode=log,
		xmin=0.2,
		xmax=500,
		ymax=100,
		ylabel near ticks,
		yticklabel pos=right,
		]
		\addplot[very thick,loosely dashed,color=cycle1] table [x=time, y=perc] {data/ppi-new.dat};
  	\end{axis}
        \begin{axis}[
	    scale only axis,
        xlabel={Time (seconds, log axis)},
		ylabel={Micro-F1},
		ymin=0.21,
		ymax=0.25,
    	xmode=log,
		minor tick num=5,
		ymajorgrids,
		yminorgrids,
        mark repeat=3,
		height=3cm,
		xmin=0.2,
		xmax=500,
		width=0.7\linewidth,
        legend style={at={(0.5,1.025)},anchor=south},
        legend entries={FREDE F1, FREDE \%n},
        legend columns=2,
        ]
	\addlegendimage{ultra thick, draw=cycle1};
	\addlegendimage{very thick, dashed, color=cycle1};
		\addplot[ultra thick,color=cycle1] table [x=time, y=micro2] {data/ppi-new.dat};
\addplot [cycle3, mark = *, nodes near coords=SVD,every node near coord/.style={anchor=90}] coordinates {( 250.83, 0.2366)};
		\addplot [cycle5, mark = square*, nodes near coords=ProNE,every node near coord/.style={anchor=150}] coordinates {( 0.51, 0.2429)};
\addplot [cycle6, mark = square*, nodes near coords=DeepWalk,every node near coord/.style={anchor=204}] coordinates {( 49.79, 0.2134)};
		\addplot [cycle7, mark = oplus*, nodes near coords=NetSMF,every node near coord/.style={anchor=215}] coordinates {( 99.45, 0.2386)};
		\addplot [cycle2!70!black, mark = triangle*, nodes near coords=STRAP,every node near coord/.style={anchor=160}] coordinates {( 20.76, 0.2347)};
		\addplot [cycle9, mark = diamond*, nodes near coords=VERSE,every node near coord/.style={anchor=-30}] coordinates {( 38.57, 0.2164)};
  	        \draw[thick, dotted, color=black] ({rel axis cs:0,0} -| {axis cs:0.610201456505178,0}) -- ({rel axis cs:0,1} -| {axis cs:0.610201456505178,0});
\end{axis}
        \end{tikzpicture}
	\vspace*{-4mm}
	\caption{\ourwork{} \pkv{scalably} produces an embedding \emph{at any time}; \pkr{at the dotted black line, it outperforms all contenders, including the SVD of \pkv{a PPR-like} similarity matrix,
after processing \pkr{about 20\%} of matrix rows}.
(\dppi data)}\label{fig:firstpage}
\vspace{-6mm}
\end{figure} 

\section{Introduction}\label{sec:introduction}

Low-dimensional representations, or \emph{embeddings}, of graph's nodes provide a multi-purpose tool for performing \pkv{data science} tasks such as community detection, link prediction, and node classification. Neural embeddings~\cite{perozzi2014, grover2016, tang2015, tsitsulin2018www}, computed by unsupervised representation learning over \emph{nonlinear} transformations, outperform their linear counterparts~\cite{ou2016, zhang2018} in \pkv{task performance}, and achieve scalability \pkv{via sampling a node similarity matrix, such as Personalized PageRank (PPR); however, such neural methods lack theoretically grounded error guarantees with respect to their objectives. The theoretically most well-grounded state-of-the-art method,} \netmf~\cite{qiu2018}, performs Singular Value Decomposition (SVD) on \pkn{a} dense matrix of nonlinear node similarities, and \pkn{achieves the} global optimum of its objective \pkn{by virtue of the properties of SVD}.

However, \pkn{this} \emph{optimality} comes \pkv{to the detriment} of \emph{scalability}, as \netmf needs to precompute the similarity matrix and store it in memory at cost quadratic in the number of nodes. An ideal method should achieve both \pkv{quality and scalability}.

In this paper, we propose \ourwork, \pkv{the first, to our knowledge,} \pkn{linear-space} algorithm that produces embeddings \pkn{with \emph{quality guarantees} \pkv{from a nonlinear transform}. We observe that factorization-based \pko{embeddings effectively strive to preserve the covariance of a similarity matrix},} and that \pk{a few nodes acting as oracles approximate the distances among all nodes with guarantees~\cite{thorup2005approximate}. Given these observations, we adapt a \pkv{covariance-preserving} matrix sketching algorithm, \emph{Frequent Directions} (FD)~\cite{liberty2013,ghashami2016journal}, to produce a graph embedding by factorizing, on a per-row basis, a \pkv{PPR-like node similarity matrix derived by interpreting a state-the-art neural embedding, \verseemb~\cite{tsitsulin2018www}, as matrix factorization.}} \ourwork~\pki{can be distributed, as it} inherits the \emph{mergeability} property of FD: two embeddings can be computed independently on different node sets and merged to a single embedding, with quality guarantees \pki{that hold \emph{anytime}~\cite{zilberstein96}, even} after accessing a subset of similarity matrix rows. Figure~\ref{fig:firstpage} shows that {\ourwork} \pkr{outperforms state-of-the-art methods and SVD in a node classification task after processing about 20\% of similarity matrix rows representing graph nodes.}

We summarize our contributions as follows:
\begin{enumerate}
	\item \pkn{we interpret \dmv{a} state-of-the-art graph embedding method, \verseemb, \pkv{as factorizing a transformed PPR similarity matrix;}}
	\item \pkn{we propose \ourwork, an \emph{anytime} graph embedding algorithm that minimizes covariance error \pkv{on that PPR-like matrix} via \emph{sketching}, \pkv{with space complexity linear in the number of nodes and} time linear in the number of processed rows};
	\item in a thorough experimental evaluation \dmv{with real graphs} we confirm that \ourwork is competitive against the state of the art \dmv{and scales to large networks}.
\end{enumerate}
 \section{Preliminaries and Related Work}\label{sec:related-work}

\pkv{Our work builds on the know-how of matrix sketching to derive scalable, anytime graph embeddings for practical data science tasks. Here, we outline previous work on graph embeddings and the fundamentals of matrix sketching.}

\subsection{Problem setting}\label{ssec:problem-setting}

A \emph{graph} is a pair ${G=(V,E)}$ with $n$ vertices ${V = (v_1, \ldots, v_n)}$, $|V|=n$, and edges $E \subseteq V\times V$, $|E|=m$, represented by an \emph{adjacency matrix} $\adj$ for which $\adj_{ij} = 1$ if $(i, j)\in E$ is an edge between node $i$ and node $j$, otherwise $\adj_{ij} = 0$. $\dmat$ is the diagonal matrix with the degree of node $i$ as entry $\dmat_{ii} = \sum_{j = 1}^n \adj_{ij}$. The \emph{normalized adjacency matrix}, $\pmat = \dmat^{-1}\adj$, represents the transition probability from a node to any of its neighbors. We represent interactions among nodes with a \emph{similarity matrix} $\smat\in\reals^{n\times n}$~\cite{tsitsulin2018www,ou2016,zhang2018}. The row~$i$ of $\amat$ is denoted as~$\adj_i$. \pkv{The embedding problem is to find} an~$n\!\times\!d$ matrix $\emb$ that retains most information in~$\smat$.

\newcommand{\fa}{factorization\xspace}
\newcommand{\nr}{neural\xspace}

\begin{table*}[!th]
\setlength{\belowrulesep}{0.1pt}
\setlength{\aboverulesep}{0.1pt}
\small
    \setlength{\tabcolsep}{1pt}
    \newcolumntype{C}{>{\centering\arraybackslash}X}
    \newcolumntype{T}{>{\hsize=.55\hsize}X}
    \newcolumntype{S}{>{\hsize=.55\hsize}C}
    \begin{center}
    {
\begin{tabularx}{\linewidth}{p{.7cm}p{1.4cm}*{7}{S}SC}
    \multicolumn{2}{c}{}  & \multicolumn{4}{@{}c@{}}{\textbf{Solution}} & \multicolumn{3}{@{}c@{}}{\textbf{Computation}} & \multicolumn{2}{@{}c@{}}{\textbf{Complexity}}\\
    \cmidrule(lr){3-6}\cmidrule(lr){7-9}\cmidrule(lr){10-11}
    \multicolumn{1}{l}{\emph{type}} &
    \multicolumn{1}{l}{\emph{method}} & Nonlinear & \mbox{Closed-form} & \mbox{Error-bounded} & Versatile & Frugal & Anytime & Mergeable & Space & Time\\ 
    \midrule
\parbox[t]{2mm}{\multirow{4}{*}{\rotatebox[origin=c]{90}{\nr}}} & \deepwalk &  \multirow{2}{*}{\cmark} & \multirow{2}{*}{\xmark} & \multirow{2}{*}{\xmark} & \multirow{2}{*}{\xmark} & \cmark & \multirow{2}{*}{\xmark} & \multirow{2}{*}{\xmark} & $\bigO(dn)$ & $\bigO(dn\log n)$ \\
& \nodetovec & & & & & \xmark & & & $\bigO(n^3)$ & $\bigO(dnb)$ \\& \lineemb &  \cmark & \xmark & \xmark & \xmark & \cmark & \xmark & \xmark & $\bigO(dn)$ & $\bigO(dnb)$ \\
     & \verseemb  & \cmark & \xmark & \xmark & \cmark & \cmark & \xmark & \xmark & $\bigO(dn)$ & $\bigO(dnb)$ \\
    \midrule
    \parbox[t]{2mm}{\multirow{8}{*}{\rotatebox[origin=c]{90}{\fa}}} & \hoppe &  \xmark & \cmark & \cmark & \cmark & \cmark & \xmark & \xmark & $\bigO(dn)$ & $\bigO(d^2m)$ \\
    & \arope &  \xmark & \cmark & \xmark & \cmark & \cmark & \xmark & \xmark & $\bigO(dn)$ & \mbox{$\bigO(dm\!+\!d^2n)$} \\
    & \appr  & \multirow{2}{*}{\xmark} & \cmark & \multirow{2}{*}{\xmark} & \multirow{2}{*}{\xmark} & \multirow{2}{*}{\cmark} & \multirow{2}{*}{\xmark} & \multirow{2}{*}{\xmark} & \multirow{2}{*}{$\bigO(dn)$} & \multirow{2}{*}{\mbox{$\bigO((dm + d^2n)\log n)$}} \\
    & \nrp  & & \xmark & & & & & & & \\
    & \netmf &  \multirow{2}{*}{\cmark} & \multirow{2}{*}{\cmark} & \cmark & \multirow{2}{*}{\cmark} & \multirow{2}{*}{\xmark} & \multirow{2}{*}{\xmark} & \multirow{2}{*}{\xmark} & $\bigO(n^2)$ & $\bigO(dn^2)$ \\
    & \netsmf & & & \xmark & & & & & \mbox{$\bigO(n^2)$} & \mbox{$\bigO(n^2)$} \\ & \mbox{\pkr{\strap}}  & \cmark & \xmark & \xmark & \cmark & \cmark & \xmark & \xmark & \mbox{$\bigO(\nicefrac{n}{\epsilon})$} & \mbox{$\bigO(d^2n + \nicefrac{m}{\epsilon})$} \\
     & \mbox{\pkr{\prone}} & \cmark & \xmark & \xmark & \xmark & \cmark & \xmark & \xmark & \mbox{$\bigO(dn)$} & \mbox{$\bigO(d^2n+Tm)$} \\
    \midrule
    clust & \mbox{\pkr{\louvain}} & \cmark & \xmark & \xmark & \xmark & \cmark & \xmark & \xmark & \mbox{$\bigO(m \log n)$} & \mbox{$\bigO(m \log n + dn\log n)$} \\
    \midrule
    skch & \mbox{\pkr{\nodesketch}} &  \cmark & \xmark & \xmark & \xmark & \xmark & \xmark & \xmark & \mbox{$\bigO(n^2)$} & \mbox{$\bigO(n^2)$} \\
    \midrule
    \belowrulesepcolor{cycle2!10!white}
    \rowcolor{cycle2!10!white} skch & \multicolumn{1}{l}{\ourwork~(ours)} &  \cmark & \cmark & \cmark & \cmark & \cmark & \cmark & \cmark & $\bigO(dn)$ & $\bigO(dn^2)$ \\
    \aboverulesepcolor{cycle2!10!white}
    \bottomrule
\end{tabularx}
    }
\end{center}
\caption{Fulfilled (\cmark) and missing (\xmark) desiderata of related work; complexities in terms of number of nodes $n$ and edges $m$, dimensionality $d$, context size $T$, and number of negative samples $b$; \pkv{we assume a sparse graph} \pkr{with average degree $\bigO(d)$}.
}\label{tbl:relatedwork}
\vspace{-5mm}
\end{table*}

\subsection{Neural embeddings}

\pkv{Initial works on graph embeddings relied on a neural network to produce vector representations of a graph's nodes;} \deepwalk~\cite{perozzi2014} \pkv{first transferred such methods} from words~\cite{mikolov2013,levy2014} to graphs, \pki{utilizing} a corpus of random walks. \lineemb~\cite{tang2015} extended DeepWalk by exploiting graph edges rather than walks; \nodetovec~\cite{grover2016} customized random walk generation; \pk{and \verseemb~\cite{tsitsulin2018www} generalized \pkv{this approach to a method that preserves} any similarity measure among nodes, with Personalized PageRank (PPR)~\cite{page1999} as the default option. Such neural embeddings \pkv{leverage paths around a node, reach scalability via sampling, and} \pkn{provide no closed-form solution and no quality guarantees;} we call them \dmv{\emph{positional}} embeddings. In another neural approach,} \dmv{\emph{structural} embeddings~\cite{ahmed2020role, ribeiro2017struc2vec, rossi2020structural} leverage \pkv{complex graph structural patterns} to improve quality \pkv{at the expense of} scalability; however, positional embeddings outperform \pkv{their structural counterparts} in both quality and scalability. For these reasons, we exclude structural embeddings from our \pkv{discussion}. }

\subsection{Factorization-based embeddings}

\pkv{Other works cast the problem of embedding a graph's nodes as one of exact or approximate factorization of the node similarity matrix, which is meant to minimize} the \pki{\emph{reconstruction error}~\cite{qiu2018}}:

\begin{definition}[Reconstruction error]\label{def:frobenius-error}
The \emph{reconstruction error} between matrices $\smat$ and $\apsim$ is the Frobenius norm of the  difference among the $\smat$ and $\apsim$, i.e., ${\lVert \smat - \apsim\rVert_F^2 = \sqrt{\sum_{i=1}^{n}\sum_{j=1}^n (\smat_{ij} - \apsim_{ij})^2}}$.
\end{definition}

In the case of \emph{symmetric} $\smat$, there exists an eigendecomposition ${\smat = \umat \Lambda \umat^{\top}}$, and the optimal rank-$k$ approximation \pkv{of $\smat$} is ${[\smat]_k = \wmat \wmat^{\top}}$, \pki{where the matrix} ${\wmat=\umat_k\sqrt{\Lambda_k}}$ is the product \pki{between the matrix of} the first $k$ eigenvectors, $\umat_k$, and a diagonal matrix of the square roots of the first $k$ eigenvalues, $\Lambda_k$. \pkv{On the other hand,} in case $\smat$ is \emph{asymmetric}, the best rank-$k$ approximation \pki{is} obtained by the first $k$ singular vectors and values \pki{of the Singular Value Decomposition (SVD)} ${\smat = U\Sigma V^\top}$, i.e., $[\smat]_k = \umat_k \Sigma_k \vmat_k^{\top}$, where $\umat_k$ and $\vmat_k$ \pki{denotes} the first $k$ columns of $\umat$ and $\vmat$, respectively.

GraRep~\cite{cao2015} \pkv{applies SVD to factorize} a concatenation of dense log-transformed DeepWalk transition probability matrices over different numbers of steps; \pki{yet it is neither scalable, nor provides quality guarantees.} \hoppe~\cite{ou2016} overcomes the scalability drawback \pki{using} a generalized form of SVD on special similarity matrices in the form~$AB^{-1}$; it achieves optimality due to the guarantees of the Eckart--Young--Mirsky theorem, but its overall performance is hindered by \pkv{the linearity of the underlying transform}~\cite{tsitsulin2018www}. \arope~\cite{zhang2018} applies spectral filtering on symmetric similarity matrices, forfeiting \pki{any} guarantees. \pkv{\appr~\cite{yang2020} applies the randomized block Krylov SVD algorithm~\cite{musco2015} on a truncated PPR matrix; \nrp~\cite{yang2020} iteratively reweights the resulting embedding vectors by coordinate descent; this post-hoc refinement boosts the performance of \appr embeddings~\cite{yang2020}.} \pkr{\strap~\cite{yin2019scalable} applies sparse factorization on a sparse-PPR-based proximity matrix; similarly, \prone~\cite{zhang2019prone} uses sparse factorization on a weighted adjacency matrix and spectral propagation on the obtained embeddings.}

 \subsection{The neural-factorization connection}

\pkv{Recent work has established a connection between neural and factorization-based embeddings.} In \netmf~\cite{qiu2018}, Qiu et al. extended an analysis of word embeddings~\cite{levy2014} to connect matrix factorization and neural embeddings: under certain probability independence assumptions, \deepwalk, \lineemb, and \nodetovec implicitly apply SVD on dense log-transformed similarity matrices. \netmf proposes novel closed-form solutions to compute such matrices \pkn{with optimal error guarantees}. For example, \deepwalk's objective is equivalent to SVD on the dense similarity matrix 

\vspace{-3mm}
\begin{equation}\label{eq:netmf-matrix}
    \smat = \log \left( \frac{m}{bT} \big( \sum_{r=1}^{T} \pmat^r \big) \dmat^{-1} \right),
\end{equation}
\vspace{-2mm}

where $T$ is the random walk window size and $b$ is the number of negative samples~\cite{qiu2018}. The $d$-dimensional DeepWalk embedding is \pki{obtained as} $\umat_d\sqrt{\Sigma_d}$, where $\umat_d$ contains the $d$ left singular vectors and $\Sigma_d$ the first $d$ singular values. \pkn{However, this \netmf approach requires $\bigO(n^2)$ space to store $\smat$, a prohibitive complexity that hinders its application to graphs with more than 100\,000 nodes. In another attempt, \netsmf~\cite{qiu2019}, Qiu et al. sought to mitigate \netmf's scalability drawback by sparsifying the similarity matrix; however, \pki{matrix sparsification} \pkn{forfeits optimality} guarantees, causing performance deterioration for effectual sparsity levels~\cite{qiu2019}; besides, the sparsified matrix has $\bigO(Tm\log n)$ nonzeros, hence it still \pkn{yields} quadratic growth.} \subsection{Synoptic overview}

Table~\ref{tbl:relatedwork} presents previous work, \pkr{including a hierarchical\hyp{}clustering\hyp{}based heuristic, \louvain~\cite{bhowmick20louvain}, and one sketching Self\hyp{}Loop\hyp{}Augmented adjacency vectors, \nodesketch~\cite{yang19nodesketch}}, in terms of desiderata of the \emph{solution}, its \emph{computation}, and time/space requirements:

\begin{itemize}[leftmargin=0.1cm,itemindent=.3cm,labelwidth=\itemindent,labelsep=0cm,align=left]
	\item \textbf{nonlinear:} using nonlinear transforms; \pkv{\hoppe, \arope, and \appr/\nrp use linear transforms; nonlinearity is desirable, as linear dimensionality reduction methods fail to confer the advantages of their nonlinear counterparts in general~\cite{lee07book, tsitsulin2018www}}.
	\item \textbf{closed-form:} \pkv{deriving the solution via} an explicit, well-defined formula \pkv{without relying on heuristic learning components}; only \netmf{} \pkn{and \netsmf are both closed-form and nonlinear;} \pkv{\nrp loses the closed-form character of \appr due to its performance\hyp{}boosting post-hoc heuristic reweighting; such reweighting may augment any embedding with additional node degree information, yet it was only applied on \appr in~\cite{yang2020}.}
	\item \textbf{error-bounded:} \pkn{affording nontrivial, \pkv{end-to-end} error guarantees} \pkv{with respect to
	a fixed objective}; \pkv{in principle, error-bounded methods, like \hoppe and \netmf, are closed-form; the reverse is not always the case, as some closed-form methods}
	\pkv{abandon \pkn{guarantees} for sake of scalability: \arope by spectral filtering, \appr by truncating, and \netsmf by sparsifying the similarity matrix.}
	\item \textbf{versatile:} accommodating \pkv{diverse} similarity measures; \pkr{\hoppe, \arope, \verseemb, \netmf \& \netsmf, and \strap are versatile.}
\item \textbf{frugal} (space-efficient)\textbf{:} \pkn{having worst-case space complexity subquadratic in the number of nodes.}
	\item \textbf{anytime:} allowing the computation of a {\em partial} embedding \pkn{whose quality improves} as more nodes are processed.
	\item \textbf{mergeable:} \pki{allowing for a combination of} embeddings on two node subsets \pki{that retains} guarantees, \pki{hence} \pkv{enabling} distributed computation~\cite{agarwal2013}.
\end{itemize} 

\subsection{Matrix sketching}\label{ssec:sketching-review}

\pkn{SVD applied on a matrix $\mmat\in\reals^{s\times t}$ ($s$ elements, $t$ features) produces $[\mmat]_k = \umat_k \Sigma_k \vmat_k^{\top}$ that minimizes reconstruction error; from the same SVD we \pkr{can also obtain} \pkv{$\wmat= \Sigma_k\vmat_k^{\top}$}, which minimizes \emph{column covariance error}, dependent on the singular value decay of $\mmat$:}

\vspace{-1mm}
\begin{definition}[Covariance error]
	The \emph{column covariance error} is the normalized difference between the covariance matrices:
{
	$$\mathtt{ce}_k(\mmat, \wmat) = \frac{\lVert \mmat^\top\mmat - \wmat^\top\wmat\rVert_2}{\lVert \mmat - [\mmat]_k \rVert_F^2} \geq \frac{\lVert \mmat^\top\mmat - \wmat^\top\wmat\rVert_2}{\lVert\mmat\rVert_F^2} = \mathtt{ce}(\mmat, \wmat)$$
	}
\end{definition}

\pkr{\emph{Matrix sketching}~\cite{boutsidis2009, boutsidis2011, liberty2013, woodruff2014book} is an alternative to the \pkr{computationally heavy matrix reconstruction by SVD} grounded on the connection between SVD and covariance error;} it finds a low-dimensional matrix, or \emph{sketch}, $\wmat\in\reals^{d \times t}$ of~$\mmat$, \pkv{by \emph{row-wise} streaming} and \pkr{with \pkn{guarantees on} \emph{column covariance} error,} which accounts for variance loss in each dimension. The correct $k$ for the best rank $k$ approximation $[\mmat]_k$ is not known and often requires grid search. Hence, we \pko{use the \pkq{lower} bound} $\mathtt{ce}(\mmat, \wmat)$ in lieu of $\mathtt{ce}_k(\mmat, \wmat)$.

\pkv{A desirable sketch property} is \emph{mergeability}:

\vspace{-2mm}
\begin{definition}
\mpara{Mergeability.} A sketching algorithm $\mathtt{sketch}$ is \emph{mergeable} if there exists an algorithm $\mathtt{merge}$ that, applied on the $d\times t$ sketches, $\wmat_1 = \mathtt{sketch}(\mmat_1)$ and $\wmat_2 = \mathtt{sketch}(\mmat_2)$, \pkw{of two $\frac{s}{2} \times t$ matrices, $\mmat_1$, $\mmat_2$}, with ${\mathtt{ce}(\mmat_1, \wmat_1)\leq\epsilon}$ and ${\mathtt{ce}(\mmat_2, \wmat_2)\leq\epsilon}$, produces a $d \times t$ sketch $\wmat$ of \pkw{the concatenated matrix} $\mmat = [\mmat_1; \mmat_2]$, ${\wmat = \mathtt{merge}(\wmat_1,\wmat_2) = \mathtt{sketch}(\mmat)}$, that preserves the covariance error bound $\epsilon$, i.e., $\mathtt{ce}(\mmat, \wmat)\leq\epsilon$.
\end{definition}

\pkn{We now discuss some representative sketching algorithms.}

\para{Hashing.} We construct a 2-universal hash function ${h: [s] \rightarrow [d]}$ and a 4-universal hash function ${g: [s] \rightarrow \{-1, +1\}}$. \pki{Starting with a zero-valued} sketch matrix $\wmat$, each row $\mmat_{i}$ is added to the $h(i)$-th sketch matrix row with sign $g(i)$: $\wmat_{h(i)} = g(i) * \mmat_{i}$, with complexity linear in matrix size, $\bigO(st)$. In practice, random assignment of rows is used instead of a hash function. Setting ${d = \bigO (\nicefrac{t^2} {\epsilon^2})}$, hashing achieves ${\mathtt{ce} \leq \epsilon}$~\cite{woodruff2014book}. This sketch is trivially mergeable: ${\mathtt{merge} (\wmat_1, \wmat_2) = \wmat_1+\wmat_2}$.

\spara{Random Projections} are a fundamental data analysis tool~\cite{woodruff2014book}. Boutsidis~et~al.~\shortcite{boutsidis2011} propose a row-streaming matrix sketching algorithm that randomly combines rows of the input matrix. In matrix form, $\tilde{\mmat} = \mathrm{R}\mmat$, where the elements ${\mathrm{R}_{ij}}$ of the  $d \times s$ matrix $\mathrm{R}$ are uniformly from $\{-1/\sqrt{d}, 1/\sqrt{d}\}$. For each row $\mmat_i$, the algorithm samples a random vector $\rvec_i \in \reals^d$ with entries in $\{-1/\sqrt{d}, 1/\sqrt{d}\}$ and updates $\wmat = \wmat + \rvec_i\mmat_i^\top$. This sketch achieves ${\mathtt{ce}\leq\epsilon}$ with $d = \bigO(\nicefrac{t}{\epsilon^2})$, with practical performance exceeding the guarantee~\cite{li2006}, and is mergeable with ${\mathtt{merge} (\wmat_1, \wmat_2) = \wmat_1 + \wmat_2}$.

\spara{Sampling.} The Column Subset Selection Problem (CSSP)~\cite{boutsidis2009} is to select a column subset of an entire matrix. In the row-update model, a solution is found by sampling scaled rows $\mmat_i / \sqrt{dp_i}$ with probability $p_i = \Vert \mmat_i \Vert^2 / \Vert \mmat\Vert_F^2$. While the norm $\Vert \mmat\Vert_F^2$ is usually unknown in advance, the method can work with $d$ reservoir samplers, where $d$ is the sketch size. This sketch achieves ${\mathtt{ce}\leq\epsilon}$ with $d = \bigO(\nicefrac{t}{\epsilon^2})$, yet the cost of maintaining reservoir samples is non-negligible. The sketch is mergeable if we use distributed reservoir sampling.

\spara{Frequent Directions} (FD)~\cite{liberty2013}, \pknn{the current state of the art in sketching,} \pk{extends the Misra\hyp{}Gries algorithm~\shortcite{misra82} from frequent items to matrices and outperforms \pknn{other methods}~\cite{clarkson2013, boutsidis2009, boutsidis2011} in quality.} FD sketches a matrix by iteratively filling the sketch with incoming rows, performing SVD on it when it cannot add more rows, and shrinking the accumulated vectors with a low-rank SVD approximation. The complexity is $\bigO(dts)$, due to $\nicefrac{s}{d}$ iterations of computing the $\bigO(d^2t)$ SVD decomposition of a $2d \times t$ matrix $\wmat$ with $d \ll t$. This sketch achieves ${\mathtt{ce}{\leq}\epsilon}$ when ${d = \bigO(\nicefrac{t}{\epsilon})}$ and is mergeable with

\vspace{-2mm}
\begin{equation}\label{eq:merge}
{\mathtt{merge}(\wmat_1, \wmat_2) = \mathtt{FD} (\mathtt{concatenate} (\wmat_1, \wmat_2))}.
\end{equation}
\vspace{-3mm}

The table below lists the embedding dimension $d$ required to attain error bound ${\mathtt{ce}\leq\epsilon \leq 1}$ for different algorithms.

\begin{table}[!ht]
\vspace{-3mm}
    \setlength{\tabcolsep}{1pt}
    \newcolumntype{C}{>{\centering\arraybackslash}X}
    \begin{center}
{
    \begin{tabularx}{\linewidth}{p{2.25cm} C C C C}
    \toprule
    Algorithm & Hashing & RP & Sampling & FD \\
    \midrule
    Dimension $d$ & $\bigO(t^2/\epsilon^2)$ & $\bigO(t/\epsilon^2)$ & 
    $\bigO(t/\epsilon^2)$ & 
    $\bigO(t/\epsilon)$ \\
    \bottomrule
    \end{tabularx}
    }
\end{center}
\end{table} 
\pkv{We observe that, by putting the node similarity matrix $\smat$ in the role of the sketched matrix $\mmat$, we can effectively turn a sketching technique to an embedding method.} \pkn{Indeed, recent work~\cite{zhang2018billion} has adapted \pki{a sketching algorithm~\cite{vempala2005,boutsidis2011}} to graph embeddings, yet \pknn{forfeited\footnote{\pkv{In our experiments, we use a variant of~\cite{zhang2018billion} with error guarantees as a baseline.}} its error guarantees.} \pkv{We apply the know\hyp{}how of state-of-the-art matrix sketching to serve graph embedding purposes, leading to \emph{anytime graph embeddings} \pki{with} error guarantees.}}  \section{Anytime graph embeddings}\label{sec:solution}

\pkv{We observe that SVD-based graph embeddings, such as \hoppe and \netmf,} \pkn{\pko{use \pki{only} one of the two unitary matrices SVD produces,~$\umat$~and~$\vmat$. For example, NetMF returns $\wmat = \umat_{:d}\sqrt{\Sigma}_{:d}$, with $\Sigma$ truncated to $d$ singular values. \pki{Therefore,} such methods cannot reconstruct matrix $\smat$; SVD products $\umat$ and $\Sigma$ \emph{may only} reconstruct the \emph{row covariance matrix} $\smat \smat^{\top} = \umat \Sigma^2 \umat^{\top}$, as $\wmat \wmat^{\top} = \umat \Sigma^2 \umat^{\top}$, where $\wmat = \umat\Sigma$;}} \pkv{thus, such methods are better understood as implicitly} \pkn{minimizing the covariance error, \pkv{rather than the reconstruction error~\cite{qiu2018},} in relation to a} similarity matrix among graph nodes.

\pkv{Serendipitously,} \pko{\emph{sketching algorithms} aim to reconstruct the \emph{column covariance} $\smat^{\top}\smat = \vmat \Sigma^2 \vmat^{\top}$. Given this relationship, we apply a state-of-the-art \emph{matrix sketching algorithm} in lieu of SVD to construct a graph embedding in \emph{anytime} fashion, by \emph{row updates} of any partially materializable similarity matrix~$\smat$. Unfortunately, the matrix form of DeepWalk (Eq.~\ref{eq:netmf-matrix}) cannot be partially materialized.} \dmv{Next, we propose a partially materializable matrix based on Personalized PageRank (PPR), \pkv{inspired from} the \verseemb~\cite{tsitsulin2018www} similarity-based embeddings. As we show in the experiments, this choice attains good quality and time performance. \pkv{However, our method carries no prejudice with regard to the partially materializable matrix used; other choices are possible, such as, for example,} the Node-Reweighted PageRank (NRP)~\cite{yang2020}.} \pkv{Our aim is to illustrate the advantageous application of sketching for embedding purposes, while our framework supports any way of deriving the primary input matrix.}

\subsection{\pko{A row-wise computable similarity matrix}}\label{ssec:sim-matrices}

\pko{\verseemb~\cite{tsitsulin2018www} is the first similarity-based embedding method that does \emph{not} require the entire matrix as input, as it allows for efficient row-wise computation; in its default version, it uses the PPR similarity measure:}

\begin{definition}\label{def:ppr}
	Given a starting node distribution $s$, damping factor $\alpha$, and the transition probability matrix $\mathrm{P}$, the PPR vector $\ppr_{i\cdot}$ is defined by the recursive equation:
	\begin{equation}\label{eq:pprrec}
			\ppr_{i} = \alpha s + (1-\alpha)\ppr_{i}^\top \mathrm{P} \end{equation}
\end{definition}

To compute $\ppr_{i}$, we leverage the fact that the probability \pki{distribution} of a random walk with restart converges to $\ppr_{i}$ vector~\cite{page1999,bahmani2010}. Following~\cite{levy2014,qiu2018} we \pki{show} that, under mild assumptions, \verseemb with PPR similarity \pki{virtually} factorizes the $\log(\ppr)$ matrix up to an additive constant.

\begin{theorem}\label{th:verse-as-mf}
\pki{Let $\xmat$ be the matrix of \verseemb embeddings. If the terms $z_{ij} = \xvec_{i}^{\top} \xvec_{j}^{\vphantom{\top}}$ are independent, then \verseemb factorizes} the matrix $\ymat = \log(\ppr) + \log n - \log b = \xmat\xmat^{\top}$.
\end{theorem}

\begin{proof*}
Consider the \verseemb objective function for the uniform sampling distribution and \textrm{PPR} similarity:

\vspace{-3mm}
\begin{equation*}
	\mathcal{L} = \sum_{i=1}^{n} \sum_{j=1}^{n} \left[ \ppr_{ij} \log \sigma(\xvec_{i}^{\top} \xvec_{j}^{\vphantom{\top}}) + b \mathbb{E}_{j\prime \sim \mathcal{Q}_i} \log \sigma(-\xvec_{i}^{\top} \xvec_{j\prime}^{\vphantom{\top}}) \right],
\end{equation*}
\vspace{-2mm}

\noindent where $\sigma(x)=(1+e^{-x})^{-1}$ is the sigmoid, $\mathcal{Q}_i$ is the noise sample distribution, \pkw{and $b$ the number of noise samples}. Since $\ppr$ is right-stochastic and $\mathcal{Q}_i$ is uniform, i.e., $\Pr(\mathcal{Q}_i = j) = \frac{1}{n}$, we can separate the two terms as follows:

\vspace{-2mm}
$$\mathcal{L} = \sum_{i=1}^{n} \sum_{j=1}^{n} \ppr_{ij} \log \sigma (\xvec_{i}^{\top} \xvec_{j}^{\vphantom{\top}}) + \frac{b}{n} \sum_{i=1}^{n} \sum_{j\prime=1}^{n} \log \sigma(-\xvec_{i}^{\top} \xvec_{j\prime}^{\vphantom{\top}}).$$
\vspace{-1mm}

\pkn{An} individual loss term for vertices $i$ and $j$ is:

\vspace{-2mm}
$$\mathcal{L}_{ij} = \ppr_{ij} \log \sigma (\xvec_{i}^{\top} \xvec_{j}^{\vphantom{\top}}) + \frac{b}{n} \log \sigma(-\xvec_{i}^{\top} \xvec_j^{\vphantom{\top}}).$$
\vspace{-1mm}

We substitute $z_{ij} = \xvec_{i}^{\top} \xvec_{j}^{\vphantom{\top}}$, use our independence assumption, and solve for $\frac{\partial \mathcal{L}_{ij}}{\partial z_{ij}} = \ppr_{ij} \sigma(- z_{ij}) - \frac{b}{n} \sigma(z_{ij}) = 0$ to get $z_{ij} = \log \frac{n \cdot \ppr_{ij}}{b}$, hence $\xmat\xmat^{\top} = \log(\ppr) + \log n - \log b = \ymat.$
\end{proof*}

\pk{Even though this solution is algebraically impossible, as it implies approximating a non-symmetric matrix by a symmetric one, it provides a matrix whose covariance we can sketch.} \pkr{Similarly, \strap~\cite{yin2019scalable} applies Sparse Randomized SVD to the logarithm of a similarity matrix based on Sparse PPR.}

\subsection{\ourwork algorithm}\label{ssec:our-algorithm}

\pko{Since the matrix $\ymat = \xmat\xmat^{\top}$ has equal row and column ranks, we rewrite the decomposition commutatively, as $\ymat = \log(\ppr) + \log n - \log b = \xmat^{\top}\xmat$. We keep the bias parameter $b$ equal to $1$, as in \netmf, and apply Frequent Directions (Section~\ref{ssec:sketching-review}) to obtain a $d \times n$ sketch-based embedding $\wmat$} by processing \pkv{rows of $\ymat$}. Algorithm~\ref{alg:ourwork} presents the details of \ourwork and Figure~\ref{fig:workflow} shows its workflow; \pko{it computes rows of the $\ppr$ matrix, \pkv{and hence of the transformed $\ymat$,} by sampling \pkr{or power iterations},} applies the SVD-based Frequent Directions sketching process periodically with each $d$ rows it processes (Lines 8--12), \pkn{and returns embeddings with guarantees at any time (Lines 14--15)}. \pknn{We keep track of singular values in $\hat{\Sigma}$ alongside the sketch so as to avoid performing SVD} \pko{upon a request for output;} \pko{as in~\cite{qiu2018}, \pkv{we multiply by $\sqrt{\hat{\Sigma}}$ at output time (Line~15), whereas a covariance\hyp{}oriented sketcher would use~$\hat{\Sigma}$}.} The time to process \emph{all} $n$ nodes \pkv{with $\bigO(\sfrac{n}{d})$ SVD iterations costing $\bigO(d^2n)$} is $\bigO(dn^2)$. 

\pkn{\emph{Sketch-based embeddings} inherit the \pki{covariance error bounds} of sketching (Section~\ref{ssec:sketching-review}),} which \pkn{hold \emph{anytime}, even after processing only an arbitrary subset of rows.} Thus, \pk{\ourwork embeddings \pki{inherit the \emph{anytime error guarantees} of Frequent Directions, which are valid after \pkw{materializing} only part of the similarity matrix,}} and \pko{superior to those of \pknn{other sketch-based embeddings}}; \pkn{it achieves ${\mathtt{ce} \leq \epsilon}$ \pkn{on the submatrix $\smat_{[s]}$ built from any size-$s$ subset of processed rows (nodes) when ${d = \bigO(\nicefrac{n}{\epsilon})}$~\cite{ghashami2016journal}, independently of $s$}.} \pkn{In Section~\ref{ssec:anytime classification} we show that~\ourwork outperforms other sketch-based embeddings in anytime node classification.}

\begin{figure}
\centering
\begin{tikzpicture}
\node (caption1) at (1.05, 1.3) {\textbf{Insert $\mathrm{PPR}(v, \cdot)$ into $\wmat$}};
\foreach \nodename/\x/\y/\nc in {
    0/0/0/100,
    1/-0.75/0.25/30,
    2/-0.5/1/50,
    3/-1/-0.5/15,
    4/-0.55/-0.65/75,
    5/0/-1/35,
    6/0.5/0.5/45,
    7/0.35/-0.75/30,
    8/0/0.35/30}
{
    \node (\nodename) at (\x,\y) [shape=circle,inner sep=2pt,draw,thick,fill=cycle5!\nc!white] {};
}
\node (v) at (0.2, -0.15) {$v$};
\begin{pgfonlayer}{background}
\path
\foreach \startnode/\endnode in {
    0/1, 0/3, 0/4, 0/5, 0/6, 0/7, 0/8,
    3/4, 4/5, 2/6, 8/2, 8/6, 2/1}
{
(\startnode) edge[-,thick] node {} (\endnode)
};
\end{pgfonlayer}
\draw [->, ultra thick] (0.15, 0) -- (1.8, 0);
\draw [ultra thick] (1.875,-1.35) rectangle (2.825,1.075);
\foreach \y in {-1.2,-0.9,...,1.2} {
      \foreach \x in {2.05,2.35,...,2.7} {
          \pgfmathparse{0.8*rnd+0.1}
          \definecolor{MyColor}{rgb}{\pgfmathresult,\pgfmathresult,\pgfmathresult}
          \node[fill=MyColor,outer sep=0pt,inner sep=4pt,anchor=center] at (\x,\y) {};
      }
  }
\foreach \y/\nc in {
    -1.2/45,
    -0.9/30,
    -0.6/75,
    -0.3/100,
    0/30,
    0.3/45,
    0.6/50,
    0.9/45}
{
    \node[fill=cycle5!\nc!white,outer sep=0pt,inner sep=4pt,anchor=center] at (2.65,\y) {};
}
\node (caption1) at (4.75, 1.3) {\textbf{Compress $\wmat$, update $\hat{\Sigma}$}};
\draw [->, ultra thick] (2.95, 0) -- (4, 0);
\draw [ultra thick] (4.075,-1.35) rectangle (5.025,1.075);
\foreach \y in {-1.2,-0.9,...,1.2} {
      \foreach \x in {4.25,4.55} {
          \pgfmathparse{0.8*rnd+0.1}
          \definecolor{MyColor}{rgb}{\pgfmathresult,\pgfmathresult,\pgfmathresult}
          \node[fill=MyColor,outer sep=0pt,inner sep=4pt,anchor=center] at (\x,\y) {};
      }
  }
\node (comma) at (5.2, -0.15) {\textbf{,}};
\draw [ultra thick] (5.5,-0.35) rectangle (6.4,0.55);
\node[fill=black!70!white,outer sep=0pt,inner sep=4pt,anchor=center] at (5.65,0.4) {};
\node[fill=black!30!white,outer sep=0pt,inner sep=4pt,anchor=center] at (5.95,0.1) {};
\draw[->,very thick,decorate, decoration={snake, segment length=5mm, amplitude=0.7mm}] (5.25,-0.4) -- (6.5,-1.2);
\node [rotate=-35,anchor=north] at (5.85,-0.8) {anytime};
\end{tikzpicture}
\caption{\ourwork samples $\log(\ppr)$ rows, periodically compresses the derived sketch and gets singular values by SVD, and yields an embedding with error guarantees at any time.}\label{fig:workflow}
\vspace{-4mm}
\end{figure} 
\vspace{-2mm}
\begin{algorithm}[!h]
\begin{algorithmic}[1]
    \Function{\ourwork}{$G, n, d$} \State{$\wmat \gets \texttt{zeros}(2d, n)$} \Comment{} {all zeros matrix $\wmat \in \mathrm{R}^{2d \times n}$}
        \State{$\hat{\Sigma} \gets \mathrm{I}(2d)$} \Comment{} {diagonal identity matrix $\hat{\Sigma} \in \mathrm{R}^{2d \times 2d}$}
        \For{$v \in V$}
         \State{$x \gets \mathtt{PersonalizedPageRank}(v)$}
         \State{$y \gets \log x+\log n$} \Comment \pkv{PPR-like} similarity row
         \State{Insert $y$ into the last zero valued row of $\wmat$}
         \If{$\wmat$ has no zero valued rows}
              \State{$\umat, \Sigma, \vmat^\top \leftarrow \mathtt{SVD}(\hat{\Sigma}\wmat), \sigma \leftarrow \Sigma_{d,d}$}
\State{$\hat{\Sigma}_{:d} \gets \sqrt{ \max({\textstyle\Sigma^2_{:d}} - \sigma^2\mathrm{I}_d, 0)}$}
              \Comment \pkw{set $d^\mathrm{th}$ row of $\hat{\Sigma}$ to $0$}
              \State{$\hat{\Sigma}_{d:}\gets \mathrm{I}_d$}
              \Comment \pkw{set last $d$ entries of $\hat{\Sigma}$ to $1$}
              \State{$\wmat_{:d} \gets \vmat^\top_{:d}, \wmat_{d:} \gets \mathbf{0}_{d \times n}$}
              \Comment zero last $d$ rows of $\wmat$
         \EndIf
       \EndFor
        \State{\Return{$\hat{\Sigma}, \wmat_{:d}$}}
    \EndFunction{}
 	\Function{GetEmbedding}{$k\leq d$} \Comment{} Anytime
\State{\Return{$\sqrt{\hat{\Sigma}} \wmat_{:k}$}} \Comment{} first $k$ rows
    \EndFunction{}
    \end{algorithmic}
    \caption{\ourwork algorithm}\label{alg:ourwork}
\end{algorithm}

\subsection{Parallelization and distribution}\label{ssec:distributed-version}

\atv{}\pkv{The steps of Algorithm~\ref{alg:ourwork} are parallelizable. Line~5 could employ approximate PPR~\cite{yoon2018tpa,yang2020}, and Line~9 efficient SVD calculations~\cite{holmes2009quic}. Such speedups trade quality for scalability. Furthermore, \ourwork can be efficiently \emph{distributed} across machines for the sake of scalability, with very small communication overhead and \emph{preserving} its quality guarantees. This appealing characteristic, unique among related works on embeddings, follows from the mergeability property that \ourwork inherits from Frequent Directions. In each machine~$m$, we may create a partial embedding matrix $W$ based on the subset of the nodes available to~$m$, and then \texttt{merge} partial embeddings from $t$ servers, i.e., iteratively sketch their concatenations by Equation~\ref{eq:merge} in hierarchical fashion, incurring a $\log_2 t$ time complexity factor.} \section{Experiments}\label{sec:experiments}

\pkv{The primary advantage of \ourwork is its \emph{anytime} character, i.e., its ability to derive embeddings by processing only a fraction of similarity matrix rows. On that front, it may only be compared against other sketch-based embeddings. Here, we also compare \ourwork on qualitative performance in data science tasks against other graph embeddings to corroborate its practical impact.}

\subsection{Compared methods}

We evaluate \ourwork against three sketching baselines, exact matrix factorization by SVD, \pkr{and all methods in Table~\ref{tbl:relatedwork} bar those that (i)~use linear transforms, which previous work~\cite{tsitsulin2018www} has established underperform nonlinear ones (i.e., \hoppe, \arope, \appr, \nrp), (ii)~require heavy hyperparameter tuning (i.e., \nodetovec and \nodesketch ), or (iii)~underperform \deepwalk (i.e., \lineemb)}:

\begin{itemize}[leftmargin=0cm,itemindent=.4cm,labelwidth=\itemindent,labelsep=0cm,align=left]
    \item \pkv{Sketching baselines}, i.e., \textbf{Hashing}, \textbf{Random Projections} and \textbf{Sampling} (Section~\ref{ssec:sketching-review}), compute the sketch \pko{and filter singular values as in \ourwork}. \pkn{Our \textbf{Random Projections} baseline is a refined variant of~\cite{zhang2018billion}, substituting a crude higher-order matrix approximation with the row-update sketching algorithm applied on the transformed PPR matrix}, and \textbf{hashing} is a variant of~\cite{postavaru20}.
    \item \pkv{SVD} \pkv{is the \textbf{exact SVD decomposition} of the nonlinearly tranformed PPR matrix $\ymat$ with the same parameters as in \ourwork, against which we were able to compare on the three smallest datasets.}
	\item \deepwalk\footnote{\url{https://github.com/xgfs/deepwalk-c}}~\cite{perozzi2014} learns an embedding by sampling fixed-length random walks from each node and applying word2vec-based learning on those walks; \pkv{despite intensive research on graph embeddings, \deepwalk remains} \pko{competitive when used with time-tested default parameters~\cite{tsitsulin2018www}}: walk length $t\!=\!80$, number of walks per node $\gamma\!=\!80$, and window size $T\!=\!10$; \pko{we use these values}. 
	\item \verseemb\footnote{\url{https://github.com/xgfs/verse}}
	~\cite{tsitsulin2018www} \pkn{trains} a single-layer neural network to learn the PPR similarity measure via sampling, with default parameters $\alpha=0.85$ and {$\texttt{nsamples}=10^6$}.
    \item \netmf\footnote{Code in the supplementary material.}~\cite{qiu2018} performs SVD on the closed-form \deepwalk matrix. We use the optimal method, \netmf-small; as it is not scalable, we evaluate it on our three smallest datasets, using the same parameters as in \deepwalk, and bias $b=1$ as in the original paper. \item \pkr{\netsmf
    \footnote{\url{https://github.com/xptree/NetSMF}}
    ~\cite{qiu2018}} \dmr{sparsifies the \netmf similarity matrix to attain scalability forfeiting optimality; we run it with default $M = 10^3$.}
    \item \pkr{\louvain
    \footnote{\url{https://github.com/maxdan94/LouvainNE}}
    ~\cite{bhowmick20louvain}} \dmr{learns embeddings in a hierarchical fashion using  the Louvain hierarchical clustering method; the final node embeddings are a concatenations of cluster embeddings, with default parameters parameters $\alpha=0.01$ and no restriction on $h_{max}$, the maximum number of levels.}
    \item \pkr{\strap
    \footnote{\url{https://github.com/yinyuan1227/STRAP-git}}
    ~\cite{yin2019scalable}} \dmr{obtains embeddings through sparse factorization of approximate PPR vectors computed with a backward-push algorithm. We use the default setting with $\epsilon=10^{-5}$.}
    \item \pkr{\prone
    \footnote{\url{https://github.com/THUDM/ProNE}}
    ~\cite{zhang2019prone}} \dmr{applies spectral propagation on embeddings obtained by sparsely factorizing a weighted adjacency matrix. We use the default parameters $k=10, \mu=0.2,$ and $\theta=0.5$.}
\end{itemize}

\subsection{Datasets}

We \pkv{experiment on} $8$ \dmv{publicly available} real\dmv{\footnote{\url{https://github.com/xgfs/verse/tree/master/data}}\textsuperscript{,}\footnote{\url{http://leitang.net/code/social-dimension/data/flickr.mat}}} datasets.

\begin{itemize}[leftmargin=0cm,itemindent=.4cm,labelwidth=\itemindent,labelsep=0cm,align=left]
	\item \dppi~\cite{stark2006,grover2016}: a protein-protein interaction dataset, where labels represent hallmark gene sets of specific biological states.
	\item \dpos~\cite{mahoney2011,grover2016}: a word co-occurrence network built from Wikipedia data. Labels tag parts of speech induced by Stanford NLP parser.
	\item \dblogcatalog~\cite{zafarani2009,tang2009}: a social network of bloggers from the blogcatalog website. Labels represent self-identified topics of blogs.
	\item \dcocit~\cite{ms2016,tsitsulin2018www}: a paper citation graph generated from the Microsoft Academic graph, featuring papers published in 15 major data mining conferences. We use conference identifiers as labels.
	\item \dcoa~\cite{ms2016,tsitsulin2018www}: a coauthorsip graph from Microsoft Academic. We use snapshots from~2014 and~2016 for link prediction.
	\item \dvk~\cite{tsitsulin2018www}: a Russian all-encompassing social network. Labels represent user genders. We use snapshots from November~2016 and May~2017 for link prediction.
	\item \dflickr~\cite{zafarani2009,tang2009}: a photo-sharing social network, where labels represent user interests, and edges messages between users.
	\item \dyoutube~\cite{zafarani2009,tang2009}: a video-based network; labels are genres.
\end{itemize}

\begin{table}[h!]
\vspace{-3mm}
\setlength{\tabcolsep}{3.5pt}
\small
\centering{
\newcolumntype{R}{>{\raggedleft\arraybackslash}X}
\newcolumntype{C}{>{\centering\arraybackslash}X}
\begin{tabularx}{\linewidth}{p{1.7cm}CCCCCR}
\multicolumn{1}{c}{} & \multicolumn{3}{c}{\textbf{Size}} & \multicolumn{2}{c}{\textbf{Statistics}} \\
\cmidrule(lr){2-4}\cmidrule(lr){5-6}
\emph{dataset} & \(|V|\) & \multicolumn{1}{c}{\(|E|\)} & \(|\mathcal{L}|\) & \mbox{Avg.\ deg.} & Density \\
    \midrule
\small\dppi{} & 4k & 77k & 50 & 19.9 &  \mbox{\(5.1 \times 10^{-3}\)} \\
\small\dpos{} & 5k & 185k & 40 & 38.7 &  \mbox{\(8.1 \times 10^{-3}\)} \\
\small\dblogcatalog{} & 10k & 334k & 39 & 64.8 &  \mbox{\(6.3 \times 10^{-3}\)} \\
\small\dcocit & 44k & 195k & 15 & 8.86 &  \mbox{\(2.0 \times 10^{-4}\)} \\
\small\dcoa & 52k & 178k & --- & 6.94 &  \mbox{\(1.3 \times 10^{-4}\)} \\
\small\dvk{} & 79k & 2.7M & --- & 34.1 &  \mbox{\(8.7 \times 10^{-4}\)} \\
\small\dflickr{} & 80k & 12M & 195 & 146.55 &  \mbox{\(1.8 \times 10^{-3}\)} \\ 
\small\dyoutube{} & 1.1M & 3M & 47 & 5.25 &  \mbox{\(9.2 \times 10^{-6}\)} \\ \bottomrule
\end{tabularx}}
\caption{Dataset characteristics: number of vertices \(|V|\), number of edges \(|E|\); number of node labels \(|\mathcal{L}|\); average node degree; density defined as \(|E|/\binom{|V|}{2}\).}\label{tbl:datasets}
\vspace{-8mm}
\end{table}

\dmv{Table \ref{tbl:datasets} summarises the data characteristics}. \pki{All algorithms are implemented in Python\footnote{\url{https://github.com/xgfs/FREDE}} and ran on a} $2\!\times\!20$-core Intel E5-2698~v4 CPU machine with 384Gb RAM \pkw{and a 64Gb memory constraint}. 

\subsection{Parameter settings}

We set embedding dimension $d=128$ unless indicated otherwise. For SVD, we use the \texttt{gesdd} routine in the Intel MKL library. For classification we use LIBLINEAR~\cite{fan2008}. We repeat each experiment $10$ times and evaluate each embedding $10$ times.

\begin{figure}[ht]
\begin{tikzpicture}
    \begin{axis}[height=5cm,width=\columnwidth,title=Covariance error,ylabel={$\mathtt{ce}$}, xlabel=$d$, xmin=10, xmax=1024, xmode=log, ymode=log, samples=150,
        legend cell align=left, legend pos=south west, legend columns=3, legend style={nodes={scale=0.65, transform shape}}]
	\addplot[very thick,color=cycle1,mark=*,mark size=3pt] table [x=d, y=fd] {data/chart-homo-d-covariance.dat};\addlegendentry{\ourwork};
	\addplot[very thick,color=cycle2,mark=diamond*,mark size=3pt] table [x=d, y=svd] {data/chart-homo-d-covariance.dat};\addlegendentry{SVD};
	\addplot[very thick,color=cycle3,mark=asterisk,mark size=3pt] table [x=d, y=s] {data/chart-homo-d-covariance.dat};\addlegendentry{Sampling};
	\addplot[very thick,color=cycle5,mark=square*,mark size=3pt] table [x=d, y=rp] {data/chart-homo-d-covariance.dat};\addlegendentry{Rand.\ Proj.};
	\addplot[very thick,color=cycle4,mark=oplus,mark size=3pt] table [x=d, y=h] {data/chart-homo-d-covariance.dat};\addlegendentry{Hashing};
	\end{axis}
\end{tikzpicture}
\vspace*{-2mm}
\caption{Covariance error vs. dimensionality $d$; FREDE approaches SVD, which yields optimal covariance error.}
\label{fig:sketching-errors}
\vspace{-4mm}
\end{figure} 
\subsection{Sketching quality}\label{ssec:sketching-quality}

\pkv{As a preliminary test, we assess our choice of sketching backbone against other sketching algorithms and the optimal rank-$k$ covariance approximation obtained by SVD on the full similarity matrix, $\tilde{\smat}^{\top}\tilde{\smat} = \vmat_d \Sigma^2_d \vmat_d^{\top}$.} Figure~\ref{fig:sketching-errors} reports the covariance error $\mathtt{ce}$ on~\dppi data, vs. the dimensionality~$d$. \ourwork outperforms the other sketching algorithms (Section~\ref{ssec:sketching-review}) by at least $2$ orders of magnitude and, as~$d$ grows, it converges to the optimal SVD solution. \pkv{This result reconfirms that the advantages of Frequent Directions versus other sketching methods transfer well to the domain of graph embeddings. For the sake of completeness, we keep comparing to other sketching methods in the rest of our study, as performance may vary depending on the downstream data science task.}

\begin{figure}[!h]
\begin{tikzpicture}
	\begin{axis}[
		xlabel={Number of random walks},
		ylabel={Micro-F1},
		xmin=1000,
		xmax=1000000,
    	xmode=log,
samples=150,
		width=\columnwidth,
		height=5cm,
        legend cell align=left,
        legend pos=south east,
        legend columns=2,
        legend style={nodes={scale=0.65, transform shape}}
	]
\addplot[very thick,color=cycle1,mark=*,mark size=3pt] table [x=nwalks, y=fd] {data/chart-homo-classification-nwalks-micro.dat};\addlegendentry{\ourwork};
	\addplot[very thick,color=cycle3,mark=asterisk,mark size=3pt] table [x=nwalks, y=s] {data/chart-homo-classification-nwalks-micro.dat};\addlegendentry{Sampling};
	\addplot[very thick,color=cycle5,mark=square*,mark options={rotate=180},mark size=3pt] table [x=nwalks, y=rp] {data/chart-homo-classification-nwalks-micro.dat};\addlegendentry{Rand.\ proj.};
	\addplot[very thick,color=cycle4,mark=oplus,mark size=3pt] table [x=nwalks, y=h] {data/chart-homo-classification-nwalks-micro.dat};\addlegendentry{Hashing};
	\addplot[very thick,color=cycle2,mark=diamond*,mark size=3pt] table [x=nwalks, y=svd] {data/chart-homo-classification-nwalks-micro.dat};\addlegendentry{SVD};
  	\draw [ultra thick, dashed, draw=cycle1] (axis cs: 1024, 0.24688270236504528) -- (axis cs: 1000000, 0.24688270236504528);
  	\draw [ultra thick, dashed, draw=cycle4] (axis cs: 1024, 0.21140986017548516) -- (axis cs: 1000000, 0.21140986017548516);
  	\draw [ultra thick, dashed, draw=cycle3] (axis cs: 1024, 0.21737715199360438) -- (axis cs: 1000000, 0.21737715199360438);
  	\draw [ultra thick, dashed, draw=cycle5] (axis cs: 1024, 0.21386868167417555) -- (axis cs: 1000000, 0.21386868167417555);
  	\draw [ultra thick, dashed, draw=cycle2] (axis cs: 1024, 0.2366) -- (axis cs: 1000000, 0.2366);
	\end{axis}
	\end{tikzpicture}
	\vspace*{-2mm}
	\caption{Classification performance of sketching algorithms on PPI data wrt.\ number of walks to compute PPR.}\label{fig:homo-classification-nwalks-micro}
\vspace{-5mm}
\end{figure} 

\subsection{PPR approximation}\label{ssec:ppr-approximation}

\pkw{Figure~\ref{fig:homo-classification-nwalks-micro} shows the performance of sketching algorithms on a node classification task (predicting correct labels) vs. the number of random walks for PPR approximation. \ourwork consistently outperforms sketching baselines and reaches the exact-PPR solution with $10^6$ walks.} \dmv{This result indicates that we can achieve performance obtained using the exact PPR values in downstream tasks even without computing such PPR values with high precision.} \pkr{In the rest of our experiments, we compute PPR values by power iterations, as that is feasible with the data we use.} 

\begin{table}[!h]
\small
\begin{center}
\setlength{\tabcolsep}{1pt}
\renewcommand{\aboverulesep}{0pt}
\renewcommand{\belowrulesep}{0pt}
\newcolumntype{C}{>{\centering\arraybackslash}X}
\begin{tabularx}{\columnwidth}{XCCCCC}
\multicolumn{1}{c}{} & \multicolumn{5}{c}{\textit{labelled nodes, \%}}\\
\cmidrule{2-6}
\multicolumn{1}{l}{\emph{method}} & 10\% & 30\% & 50\% & 70\% & 90\% \\
\midrule
\deepwalk & 16.33 & 19.74 & 21.34 & 22.39 & 23.38 \\
\netmf & 18.58 & 22.01 & 23.87 & 24.65 & 25.30 \\
\netsmf & 18.45 & 22.38 & 23.86 & 24.59 & 25.56 \\
\verseemb & 16.45 & 19.89 & 21.64 & 23.08 & 23.84 \\
\louvain & 13.8&	15.91&	16.59&	16.83&	17.06 \\
\strap & 18.25 & 21.94 & 23.47 & 24.15 & 24.91 \\
\prone & 17.56 & 22.53 & 24.29 & 25.03 & \win 26.02 \\
\ourwork & \win 19.56 & \win 23.11 & \win 24.38 & \win 25.11 & 25.52 \\
\midrule
SVD & 18.31 & 22.12 & 23.66 & \win 25.03 & \win 25.78 \\
\mbox{Rand.\ Proj.} & 16.80 & 19.99 & 21.45 & 22.38 & 23.14 \\
Sampling & 16.25 & 19.55 & 20.93 & 21.85 & 22.68 \\
Hashing & 16.73 & 19.97 & 21.51 & 22.43 & 23.44 \\
\bottomrule
\end{tabularx}
\end{center}
\caption{Micro-F1 classification, \dppi data.}\label{tab:classification-ppi-micro}
\vspace{-9mm}
\end{table} \begin{table}[!h]
\small
\begin{center}
\setlength{\tabcolsep}{1pt}
\renewcommand{\aboverulesep}{0pt}
\renewcommand{\belowrulesep}{0pt}
\newcolumntype{C}{>{\centering\arraybackslash}X}
\begin{tabularx}{\columnwidth}{XCCCCC}
\multicolumn{1}{c}{} & \multicolumn{5}{c}{\textit{labelled nodes, \%}}\\
\cmidrule{2-6}
\multicolumn{1}{l}{\emph{method}} & 10\% & 30\% & 50\% & 70\% & 90\% \\
\midrule
\deepwalk & 43.42 & 47.12 & 48.96 & 49.86 & 50.18 \\
\netmf & 43.42 & 46.98 & 48.52 & 49.23 & 49.72 \\
\netsmf & 42.06 & 44.25 & 45.17 & 45.60 & 46.25 \\
\verseemb & 40.80 & 44.70 & 46.60 & 47.65 & 48.24 \\
\louvain & 41.19&	41.62&	41.72&	41.94&	41.84 \\
\strap & 47.26 & 50.85 & 52.08 & 52.70 & 53.60 \\
\prone & \win 47.41 & \win 52.18 & \win 53.84 & \win 54.57 & \win 55.10 \\
\ourwork & 46.59 & 49.23 & 50.45 & 51.02 & 51.30 \\
\midrule
SVD & 44.69 & 48.86 & 50.57 & 51.53 & 52.20 \\
\mbox{Rand.\ Proj.} & 40.24 & 43.87 & 45.65 & 46.43 & 47.18 \\
Sampling & 40.35 & 43.80 & 45.39 & 46.30 & 46.69 \\
Hashing & 40.17 & 43.88 & 45.44 & 46.35 & 46.79 \\
\bottomrule
\end{tabularx}
\end{center}
\caption{Micro-F1 classification, \dpos data.}\label{tab:classification-pos-micro}
\vspace{-8mm}
\end{table} \begin{table}[!t]
\small
\begin{center}
\setlength{\tabcolsep}{1pt}
\renewcommand{\aboverulesep}{0pt}
\renewcommand{\belowrulesep}{0pt}
\newcolumntype{C}{>{\centering\arraybackslash}X}
\begin{tabularx}{\columnwidth}{XCCCCC}
\multicolumn{1}{c}{} & \multicolumn{5}{c}{\textit{labelled nodes, \%}}\\
\cmidrule{2-6}
\multicolumn{1}{l}{\emph{method}} & 10\% & 30\% & 50\% & 70\% & 90\% \\
\midrule
\deepwalk & 36.22 & 39.84 & 41.22 & 42.06 & 42.53 \\
\netmf & 36.62 & 39.80 & 41.05 & 41.70 & 42.17 \\
\netsmf & 35.74 & 39.16 & 40.17 & 40.82 & 41.06 \\
\verseemb & 35.82 & 40.06 & 41.63 & \win 42.63 & 43.14 \\
\louvain & 18.40&	19.99&	20.68&	21.17&	21.40 \\
\strap & \win 37.48&	40.74&	41.82&	42.40&	42.88 \\ 
\prone & 36.74&	40.19&	41.27&	41.75&	41.99 \\
\ourwork & 35.69 & 38.88 & 39.98 & 40.54 & 40.75 \\
\midrule
SVD & \win 37.60 & \win 40.99 & \win 42.10 & \win 42.66 & \win 43.47 \\
\mbox{Rand.\ Proj.} & 30.82 & 34.43 & 35.81 & 36.52 & 37.16 \\
Sampling & 29.44 & 32.32 & 33.41 & 34.04 & 34.29 \\
Hashing & 30.81 & 34.36 & 35.82 & 36.65 & 37.28 \\
\bottomrule
\end{tabularx}
\end{center}
\caption{Micro-F1 classification, \dblogcatalog data.}\label{tab:classification-blog-micro}
\vspace{-7mm}
\end{table} \begin{table}[!t]
\small
\begin{center}
\setlength{\tabcolsep}{1pt}
\renewcommand{\aboverulesep}{0pt}
\renewcommand{\belowrulesep}{0pt}
\newcolumntype{C}{>{\centering\arraybackslash}X}
\begin{tabularx}{\columnwidth}{XCCCCC}
\multicolumn{1}{c}{} & \multicolumn{5}{c}{\textit{labelled nodes, \%}}\\
\cmidrule{2-6}
\multicolumn{1}{l}{\emph{method}} & 1\% & 3\% & 5\% & 7\% & 9\% \\
\midrule
\deepwalk & 37.22 & 40.34 & 41.72 & 42.59 & 43.16 \\
\netsmf & 41.07&	43.29&	44.13&	44.61&	44.99 \\
\verseemb & 38.95 & 41.20 & 42.55 & 43.41 & 44.01 \\
\louvain & 35.29&	36.56&	37.14&	37.49&	37.67 \\
\strap & 35.86&	41.93&	43.38&	43.80&	44.12 \\
\prone & 35.82&	39.54&	40.72&	41.46&	41.97 \\
\ourwork & \win 42.46 & \win 44.56 & \win 45.39 & \win 45.84 & \win 46.17 \\
\midrule
\mbox{Rand.\ Proj.} & 40.89 & 42.63 & 43.63 & 44.32 & 44.78 \\
Sampling & 40.84 & 42.97 & 43.93 & 44.49 & 44.91 \\
Hashing & 40.86 & 42.66 & 43.65 & 44.29 & 44.83 \\
\bottomrule
\end{tabularx}
\end{center}
\caption{Micro-F1 classification, \dcocit data.}\label{tab:classification-aco-micro}
\vspace{-7mm}
\end{table} \begin{table}[!t]
\small
\begin{center}
\setlength{\tabcolsep}{1pt}
\renewcommand{\aboverulesep}{0pt}
\renewcommand{\belowrulesep}{0pt}
\newcolumntype{C}{>{\centering\arraybackslash}X}
\begin{tabularx}{\columnwidth}{XCCCCC}
\multicolumn{1}{c}{} & \multicolumn{5}{c}{\textit{labelled nodes, \%}}\\
\cmidrule{2-6}
\multicolumn{1}{l}{\emph{method}} & 1\% & 3\% & 5\% & 7\% & 9\% \\
\midrule
\deepwalk & \win 32.39 & \win 36.02 & \win 37.41 & \win 38.15 & \win 38.70 \\
\verseemb & 30.08 & 34.22 & 36.06 & 37.11 & 37.83 \\
\louvain & 22.19&	22.60&	22.73&	22.89&	23.03 \\
\strap & 30.13&	34.26&	35.75&	36.60&	37.14 \\
\prone & 30.90&	35.11&	36.63&	37.47&	38.04 \\
\ourwork & 30.90 & 32.98 & 33.86 & 34.48 & 34.88 \\
\midrule
\mbox{Rand.\ Proj.} & 28.92 & 32.21 & 33.82 & 34.76 & 35.49 \\
Sampling & 28.46 & 30.97 & 32.08 & 32.75 & 33.24 \\
Hashing & 29.07 & 32.23 & 33.77 & 34.75 & 35.48 \\
\bottomrule
\end{tabularx}
\end{center}
\caption{Micro-F1 classification, \dflickr{} data.}\label{tab:classification-flickr-micro}
\vspace{-8mm}
\end{table} \begin{table}[!t]
\small
\begin{center}
\setlength{\tabcolsep}{1pt}
\renewcommand{\aboverulesep}{0pt}
\renewcommand{\belowrulesep}{0pt}
\newcolumntype{C}{>{\centering\arraybackslash}X}
\begin{tabularx}{\columnwidth}{XCCCCC}
\multicolumn{1}{c}{} & \multicolumn{5}{c}{\textit{labelled nodes, \%}}\\
\cmidrule{2-6}
\multicolumn{1}{l}{\emph{method}} & 1\% & 3\% & 5\% & 7\% & 9\% \\
\midrule
\deepwalk &  37.96 &  40.54 &  41.75 &  42.60 &  43.37 \\
\verseemb &  38.04 &  40.50 &  41.72 &  42.59 &  43.33 \\
\louvain & 31.41&	32.28&	32.65&	33.00&	33.29 \\
\strap & 32.82&	38.02&	40.03&	41.13&	41.75 \\
\prone & \win 38.40&\win 42.20&\win 43.09&\win 43.69&\win 44.04 \\
\ourwork & 34.51 & 37.37 & 38.78 & 39.40 & 39.95 \\
\midrule
\mbox{Rand.\ Proj.} & 33.88 & 36.10 & 37.23 & 37.94 & 38.38 \\
Sampling & 33.97 & 35.66 & 36.37 & 37.19 & 37.71 \\
Hashing & 32.64 & 35.64 & 36.92 & 37.46 & 38.13 \\
\bottomrule
\end{tabularx}
\end{center}
\caption{Micro-F1 classification, \dyoutube data.}\label{tab:classification-yt-micro}
\vspace{-5mm}
\end{table}  

\begin{figure*}[ht]
	\begin{tikzpicture}
    \begin{groupplot}[group style={
                      group name=myplot,
                      group size= 5 by 1, horizontal sep=1cm},height=4cm,width=0.22\linewidth]
        \nextgroupplot[title=\dppi,ylabel={Micro-F1}, xlabel={\% of nodes visited}, xmin=0, xmax=100, ymin=0.2, samples=150, mark repeat=3]
				\addplot[very thick,color=cycle1,mark=*,mark size=2pt] table [x=perc, y=fd] {data/streaming-micro-ppi.dat};\label{lines:fd}
				\addplot[very thick,color=cycle3,mark=asterisk,mark size=2pt] table [x=perc, y=s] {data/streaming-micro-ppi.dat};\label{lines:s}
				\addplot[very thick,color=cycle5,mark=triangle*,mark options={rotate=180},mark size=2pt] table [x=perc, y=rp] {data/streaming-micro-ppi.dat};\label{lines:rp}
				\addplot[very thick,color=cycle4,mark=oplus,mark size=2pt] table [x=perc, y=h] {data/streaming-micro-ppi.dat};\label{lines:h};
  	    \draw [dashed, ultra thick, draw=cycle2] (axis cs: 0, 0.2366) -- (axis cs: 100, 0.2366);
  	    \draw [dashed, ultra thick, draw=cycle6] (axis cs: 0, 0.2164) -- (axis cs: 100, 0.2164);
  	    \draw [dashed, ultra thick, draw=cycle9] (axis cs: 0, 0.2347) -- (axis cs: 100, 0.2347);
  	    \draw [dashed, ultra thick, draw=cycle7] (axis cs: 0, 0.2386) -- (axis cs: 100, 0.2386);
  	    
		\nextgroupplot[title=\dcocit, xlabel={\% of nodes visited}, xmin=0, xmax=100, ymin=0.42, samples=150]
				\addplot[very thick,color=cycle3,mark=asterisk,mark size=2pt] table [x=perc, y=s] {data/streaming-micro-cocit.dat};
				\addplot[very thick,color=cycle5,mark=triangle*,mark options={rotate=180},mark size=2pt] table [x=perc, y=rp] {data/streaming-micro-cocit.dat};
				\addplot[very thick,color=cycle4,mark=oplus,mark size=2pt] table [x=perc, y=h] {data/streaming-micro-cocit.dat};
				\addplot[very thick,color=cycle1,mark=*,mark size=2pt] table [x=perc, y=fd] {data/streaming-micro-cocit.dat};
\draw [dashed, ultra thick, draw=cycle6] (axis cs: 0, 0.4255) -- (axis cs: 100, 0.4255);
  	    \draw [dashed, ultra thick, draw=cycle9] (axis cs: 0, 0.4338) -- (axis cs: 100, 0.4338);
  	    \draw [dashed, ultra thick, draw=cycle7] (axis cs: 0, 0.4413) -- (axis cs: 100, 0.4413);
		\nextgroupplot[title=\dflickr, xlabel={\% of nodes visited}, xmin=0, xmax=10, ymax=0.375, ymin=0.25, samples=150]
				\addplot[very thick,color=cycle3,mark=asterisk,mark size=2pt] table [x=perc, y=s] {data/streaming-micro-flickr.dat};
				\addplot[very thick,color=cycle5,mark=triangle*,mark options={rotate=180},mark size=2pt] table [x=perc, y=rp] {data/streaming-micro-flickr.dat};
				\addplot[very thick,color=cycle4,mark=oplus,mark size=2pt] table [x=perc, y=h] {data/streaming-micro-flickr.dat};
				\addplot[very thick,color=cycle1,mark=*,mark size=2pt] table [x=perc, y=fd] {data/streaming-micro-flickr.dat};
\draw [dashed, ultra thick, draw=cycle6] (axis cs: 0, 0.3606) -- (axis cs: 100, 0.3606);
  	    \draw [dashed, ultra thick, draw=cycle9] (axis cs: 0, 0.3575) -- (axis cs: 100, 0.3575);
        \nextgroupplot[title=\dblogcatalog,xlabel={\% of nodes visited}, xmin=0, xmax=100, ymin=0.3, ymax=0.425, samples=150]
				\addplot[very thick,color=cycle1,mark=*,mark size=2pt] table [x=perc, y=fd] {data/streaming-micro-blog.dat};
				\addplot[very thick,color=cycle3,mark=asterisk,mark size=2pt] table [x=perc, y=s] {data/streaming-micro-blog.dat};
				\addplot[very thick,color=cycle5,mark=triangle*,mark options={rotate=180},mark size=2pt] table [x=perc, y=rp] {data/streaming-micro-blog.dat};
				\addplot[very thick,color=cycle4,mark=oplus,mark size=2pt] table [x=perc, y=h] {data/streaming-micro-blog.dat};
  	    \draw [dashed, ultra thick, draw=cycle2] (axis cs: 0, 0.4210) -- (axis cs: 100, 0.4210);
  	    \draw [dashed, ultra thick, draw=cycle6] (axis cs: 0, 0.4163) -- (axis cs: 100, 0.4163);
  	    \draw [dashed, ultra thick, draw=cycle9] (axis cs: 0, 0.4182) -- (axis cs: 100, 0.4182);
  	    \draw [dashed, ultra thick, draw=cycle7] (axis cs: 0, 0.4017) -- (axis cs: 100, 0.4017);
        \nextgroupplot[title={\dblogcatalog \textbf{time}},xlabel={\% of nodes visited}, xmin=0, xmax=100, samples=150, ymode=log, xmode=log,
        ymax=3000,
        ylabel near ticks, yticklabel pos=right, ylabel={time, s}]
				\addplot[very thick,color=cycle1,mark=*,mark size=2pt,mark repeat=10] table [x=perc, y=fd] {data/blog-times-new-log.dat};
				\addplot[very thick,color=cycle3,mark=asterisk,mark size=2pt,mark repeat=10] table [x=perc, y=s] {data/blog-times-new-log.dat};
				\addplot[very thick,color=cycle5,mark=triangle*,mark options={rotate=180},mark size=2pt,mark repeat=10] table [x=perc, y=rp] {data/blog-times-new-log.dat};
				\addplot[very thick,color=cycle4,mark=oplus,mark size=2pt,mark repeat=10] table [x=perc, y=h] {data/blog-times-new-log.dat};
  	    \draw [dashed, ultra thick, draw=cycle2] (axis cs: 0.01, 383.7983796596527) -- (axis cs: 100, 383.7983796596527);
  	    \draw [dashed, ultra thick, draw=cycle6] (axis cs: 0.01, 148.257142117) -- (axis cs: 100, 148.257142117);
  	    \draw [dashed, ultra thick, draw=cycle9] (axis cs: 0.01, 20.76) -- (axis cs: 100, 20.76);
  	    \draw [dashed, ultra thick, draw=cycle7] (axis cs: 0.01, 924.61) -- (axis cs: 100, 924.61);
    \end{groupplot}
\path (myplot c1r1.north west|-current bounding box.north)--
      coordinate(legendpos)
      (myplot c5r1.north east|-current bounding box.north);
\matrix[
    matrix of nodes,
    anchor=south,
    draw,
    inner sep=0.2em,
    draw
  ]at([yshift=1ex]legendpos)
  {
    \ref{lines:fd} & \small\ourwork & [5pt]
    \ref{lines:rp} & \small Rand.\ Proj.& [5pt]
    \ref{lines:s} & \small Sampling & [5pt]
    \ref{lines:h} & \small Hashing & [5pt]
     \begin{tikzpicture}
     \draw[white] (0,0) -- (0.45,0);
  \draw[ultra thick, draw=cycle2] (0,0.0875) -- (0.45,0.0875);
\end{tikzpicture} & \small SVD & [5pt]
     \begin{tikzpicture}
     \draw[white] (0,0) -- (0.45,0);
  \draw[ultra thick, draw=cycle7] (0,0.0875) -- (0.45,0.0875);
\end{tikzpicture} & \small \netsmf & [5pt]
     \begin{tikzpicture}
     \draw[white] (0,0) -- (0.45,0);
  \draw[ultra thick, draw=cycle9] (0,0.0875) -- (0.45,0.0875);
\end{tikzpicture} & \small \strap & [5pt]
    \begin{tikzpicture}
     \draw[white] (0,0) -- (0.45,0);
  \draw[ultra thick, draw=cycle6] (0,0.0875) -- (0.45,0.0875);
\end{tikzpicture} & \small VERSE & [5pt]\\};
\end{tikzpicture}
\vspace{-7mm}
\caption{Classification performance of \ourwork with varying percentage of the graph as input on three datasets.}
\label{fig:micro-percgraph}
\vspace{-4mm}
\end{figure*} 
\subsection{Node classification}\label{ssec:node-classification}

Tables~\ref{tab:classification-ppi-micro}--\ref{tab:classification-yt-micro} report classification results in terms of the popular Micro-F1 measure~\cite{perozzi2014,tang2015};
Macro-F1 results are similar. \pkw{SVD is featured where it runs within 64Gb.}
\dmv{For each dataset, we repeat the experiment $10$ times and report the average.} \dmv{Surprisingly, on \dppi and \dpos, \ourwork outperforms its \pkv{exact counterpart}, SVD, and consistently supersedes its sketching counterparts across all datasets.}

\begin{table}[h!]
\small
\centering
\begin{tabularx}{\columnwidth}{XX}
Operator & Result \\
\midrule
\textsf{Average} & \((\mathbf{a} + \mathbf{b})/2\) \\
\textsf{Concat} & \([\mathbf{a}_1, \ldots, \mathbf{a}_d, \mathbf{b}_1, \ldots, \mathbf{b}_d]\) \\
\textsf{Hadamard} & \([\mathbf{a}_1 * \mathbf{b}_1, \ldots, \mathbf{a}_d * \mathbf{b}_d]\) \\
\textsf{Weighted L1} & \([|\mathbf{a}_1 - \mathbf{b}_1|, \ldots, |\mathbf{a}_d - \mathbf{b}_d|]\) \\
\textsf{Weighted L2} & \([{(\mathbf{a}_1 - \mathbf{b}_1)}^2, \ldots, {(\mathbf{a}_d - \mathbf{b}_d)}^2] \) \\
\bottomrule
\end{tabularx}
\caption{Edge embedding strategies for link prediction, nodes \(u,v \in V\) and corresponding embeddings \(\mathbf{a}, \mathbf{b} \in \mathbb{R}^d\).}\label{tab:lp-operators}
\vspace{-6mm}
\end{table} \begin{table}[!h]
\vspace{-2mm}
\small
\setlength{\tabcolsep}{1pt}
\renewcommand{\aboverulesep}{0pt}
\renewcommand{\belowrulesep}{0pt}
\begin{center}
\newcolumntype{C}{>{\centering\arraybackslash}X}
\begin{tabularx}{\columnwidth}{XCCCCC}
\multicolumn{1}{l}{\emph{method}} & \textsf{Average} & \textsf{Concat} & \textsf{Hadamard} & \textsf{L1} & \textsf{L2} \\
\midrule
\deepwalk & 68.97 & 68.43 & 66.61 & \underline{78.80} & 77.89 \\
\netsmf & 74.59&	74.24&	\underline{81.82}&	64.73&	64.57 \\
\verseemb & 79.62 & 79.25 & \underline{86.27} & 75.15 & 75.32 \\
\louvain & 67.88&	67.45&	67.85&	\underline{69.66}&	69.91 \\
\strap & 79.08&	78.72&	\underline{80.32}&	71.81&	72.05 \\
\prone & 74.15&	73.88&	74.74&	\underline{74.80}&	73.89 \\
\ourwork & 81.28 & 80.95 & \win \underline{86.83} & 81.70 & 82.37 \\
\midrule
\mbox{Rand.\ Proj.} & 80.81 & 80.54 & \win \underline{86.73} & 80.79 & 81.42 \\
Sampling & 80.98 & 80.74 & \underline{86.45} & 79.53 & 79.51 \\
Hashling & 80.84 & 80.48 & \underline{86.66} & 80.59 & 81.33 \\
\midrule
Baseline & \multicolumn{5}{c}{77.53} \\
\bottomrule
\end{tabularx}
\end{center}
\caption{Link prediction accuracy, \dcoa data.}\label{tab:link-prediction-aco}
\vspace{-6mm}
\end{table} \begin{table}[!h]
\vspace{-2mm}
\small
\setlength{\tabcolsep}{1pt}
\renewcommand{\aboverulesep}{0pt}
\renewcommand{\belowrulesep}{0pt}
\begin{center}
\newcolumntype{C}{>{\centering\arraybackslash}X}
\begin{tabularx}{\columnwidth}{XCCCCC}
\multicolumn{1}{l}{\emph{method}} & \textsf{Average} & \textsf{Concat} & \textsf{Hadamard} & \textsf{L1} & \textsf{L2} \\
\midrule
\deepwalk & 69.98 & 69.83 & 69.56 & \underline{78.42} & 77.42 \\
\netsmf & 72.63&	72.51&	68.17&	\underline{74.52}&	74.05 \\
\verseemb & 74.56 & 74.42 & \win \underline{80.94} & 77.16 & 77.47 \\
\louvain & 66.87&	66.78&	67.74&	67.42&	67.44 \\
\strap & 74.35&	74.23&	\underline{76.93} &	67.84&	66.10 \\
\prone & 71.29&	71.14&	\underline{74.92} &	73.54&	74.41 \\
\ourwork & 74.68 & 74.59 & \underline{77.63} & 74.25 & 73.60 \\
\midrule
\mbox{Rand.\ Proj.} & 74.41 & 74.27 & \underline{77.01} & 74.33 & 74.56 \\
Sampling & 74.38 & 74.27 & \underline{76.82} & 72.26 & 71.95 \\
Hashing & 74.36 & 74.27 & \underline{76.86} & 74.30 & 74.56 \\
\midrule
Baseline & \multicolumn{5}{c}{78.84} \\
\bottomrule
\end{tabularx}
\end{center}
\caption{Link prediction accuracy, \dvk data.}\label{tab:link-prediction-vk}
\vspace{-7mm} \end{table} 

\subsection{Link prediction}\label{ssec:link-prediction}

\dmv{Link prediction is the task of predicting the appearance of a link between pairs of nodes in a graph.} Tables~\ref{tab:link-prediction-aco} and~\ref{tab:link-prediction-vk} report link prediction accuracy (predicting the appearance of a link) on~\pkr{\dcoa} and~\dvk by a logistic regression classifier on features derived from embeddings by the rules in Table~\ref{tab:lp-operators}. As a baseline, we use common link prediction features (node degree, number of common neighbors, Adamic-Adar index, Jaccard coefficient, and preferential attachment). We represent absent links in the training data by negative sampling, and use 50\% of links for training and remaining 50\% for testing. \ourwork outperforms all methods on~\dcocit, and all sketching baselines on~\dvk. \dmv{Surprisingly, sketching baselines perform better than state-of-the-art graph embeddings on \dcocit.}

\subsection{Anytime classification}\label{ssec:anytime classification}

We study anytime operation (Section~\ref{ssec:our-algorithm}) \pkw{on node classification using 50\% of nodes for training and \pkw{processing} PPR rows in random order.} Figure~\ref{fig:micro-percgraph} presents results \pki{for PPR-based methods} \pkr{and \netsmf} on three datasets. \pkw{\ourwork outperforms \pkr{all others} on~PPI, as in Table~\ref{tab:classification-ppi-micro}, after processing only 1\% of similarities. \pkr{Remarkably, on \dcocit data, all sketchers outperform all other methods after processing about 3\% of nodes; their downstream performance drops thereafter, a fact indicating that more information, revealing inter-cluster connections, harms the classification outcomes. A similar effect appears for sketching baselines with {\dppi} data, yet not for \ourwork, which outperforms all others after processing less than 20\% of nodes and keeps growing thereafter.} \ourwork also performs competitively on \dflickr (we examine up to 10\% of nodes, as Random Projections was inefficient; SVD \pkr{and \netsmf} did not run within 64Gb) and \dblogcatalog, as in Tables~\ref{tab:classification-blog-micro} and~\ref{tab:classification-flickr-micro}.} \pkv{The rightmost plot in Figure~\ref{fig:micro-percgraph}} \pkw{shows runtime on \dblogcatalog; while sketchers' runtime grows linearly, those of one-off methods, \pkr{except \strap,}  stand apart.} \pkv{These results illustrate that embedding merging preserves the downstream quality; as Equation~\ref{eq:merge} shows, merging two embeddings amounts to sketching their concatenation; therefore, the sketch operation Algorithm~\ref{alg:ourwork} periodically performs with each new $d$ similarity matrix rows it processes can also be viewed as a \texttt{merge} operation.} 

\begin{figure}[h!t]
\vspace{-3mm}
\centering
\begin{tikzpicture}
    \begin{axis}[
    xlabel={Number of nodes $|V|$},
                   ymode=log,
                   xmode=log,
                   ylabel={Memory, KB},
                   ylabel style={align=center},
                   xshift=2.5em,
      yminorticks=true,
width=\columnwidth,
      height=4cm,ymajorgrids=true,
      grid style=dotted,
        legend style={at={(0.5,1.025)},anchor=south},
        legend columns=4,
        legend cell align={left},
    ]
        \addplot[color=cycle1, only marks, mark=*] table [x=v, y=FREDE] {data/mem-node.dat};\addlegendentry{\ourwork};
        \addplot[color=cycle9, only marks, mark=pentagon*] table [x=v, y=STRAP] {data/mem-node.dat};\addlegendentry{\strap};
        \addplot[color=cycle7, only marks, mark=oplus*] table [x=v, y=NetSMF] {data/mem-node.dat};\addlegendentry{\netsmf};
        \addplot[color=cycle2, only marks, mark=square*] table [x=v, y=SVD] {data/mem-node.dat};\addlegendentry{SVD};
        \addplot[color=cycle6, only marks, mark=triangle*] table [x=v, y=VERSE] {data/mem-node.dat};\addlegendentry{\verseemb};
        
        \addplot[color=cycle3, only marks, mark=asterisk] table [x=v, y=LouvainNE] {data/mem-node.dat};\addlegendentry{\louvain};
        \addplot[color=cycle4, only marks, mark=diamond*] table [x=v, y=ProNE] {data/mem-node.dat};\addlegendentry{\prone};
        \addplot[color=cycle5, only marks, mark=10-pointed star] table [x=v, y=DeepWalk] {data/mem-node.dat};\addlegendentry{\deepwalk};
        \addplot [domain=3890:1138499, no markers, densely dotted, thick, color=cycle1!50] {10^((0.8009957512*log10(x))+2.048082287)}; \addplot [domain=3890:1138499, no markers, densely dotted, thick, color=cycle9!50] {10^((0.7932131779*log10(x))+2.991496012)}; \addplot [domain=3890:1138499, no markers, densely dotted, thick, color=cycle7!50] {10^((0.9531626911*log10(x))+2.509256723)}; \addplot [domain=3890:1138499, no markers, densely dotted, thick, color=cycle2!50] {10^((1.976765128*log10(x))-1.995528074)}; \addplot [domain=3890:1138499, no markers, densely dotted, thick, color=cycle6!50] {10^((0.7604598733*log10(x))+1.086829391)}; \addplot [domain=3890:1138499, no markers, densely dotted, thick, color=cycle3!50] {10^((0.7492993206*log10(x))+0.9890453)}; \addplot [domain=3890:1138499, no markers, densely dotted, thick, color=cycle4!50] {10^((0.8529133679*log10(x))+1.817286856)}; \addplot [domain=3890:1138499, no markers, densely dotted, thick, color=cycle5!50] {10^((0.8546190266*log10(x))+0.8805520858)}; \end{axis}
\end{tikzpicture}
\vspace*{-2mm}
\caption{\dmr{Memory consumption per datasets and regression lines; SVD and NetSMF could not fit \dyoutube in 64Gb RAM.}}\label{fig:memory}
\vspace{-5mm}
\end{figure} 
\dmr{\subsection{Memory consumption}

Lastly, Figure~\ref{fig:memory} shows the memory consumption for each method and dataset, with linear regression on data sizes. All methods bar SVD and \netsmf embed the largest \dyoutube graph ($10^6$ nodes) on a 64Gb RAM commodity machine. \ourwork consumes memory comparable to that of neural methods and never exceeds that of other factorization-based embeddings.} \section{Conclusion}\label{sec:conclusion}

\pkn{Since graph embeddings implicitly aim to preserve similarity matrix covariance, row-wise sketching techniques are suited therefor. \pkw{We applied a state-of-the-art sketcher, Frequent Directions, on a matrix-factorization interpretation of \pkv{a} state-of-the-art embedding, \verseemb, to craft \ourwork: a \pkw{linear\hyp{}space} graph embedding \pkv{that allows for scalable data science operations on graph data, as well as for}} \pkw{anytime and distributed computation} \emph{with error guarantees}.} \pki{Besides its anytime character,} \pkn{\ourwork achieves almost as low covariance error as the exact SVD solution} \pkw{and stands its ground against} previous graph embeddings \pkv{even after processing as little as 10\% of similarity matrix rows; therefore, it promises significant practical impact. In the future, we plan to examine the feasibility of anytime local embeddings~\cite{postavaru20}, applications of graph embeddings in the context of graph summarization~\cite{safavi19} and anonymization~\cite{xue12}, and augment \ourwork using recent enhancements on Frequent Directions~\cite{huang2018}.} 
\balance
\bibliographystyle{ACM-Reference-Format}
\bibliography{bibliography} 

\end{document}